# Learning in Sinusoidal Spaces with Physics-Informed Neural Networks

Jian Cheng Wong, Chinchun Ooi, Abhishek Gupta, *Senior Member, IEEE*, and Yew-Soon Ong, *Fellow, IEEE*

*Abstract*— A physics-informed neural network (PINN) uses physics-augmented loss functions, e.g., incorporating the residual term from governing partial differential equations (PDEs), to ensure its output is consistent with fundamental physics laws. However, it turns out to be difficult to train an accurate PINN model for many problems in practice. In this paper, we present a novel perspective of the merits of learning in sinusoidal spaces with PINNs. By analyzing behavior at model initialization, we first show that a PINN of increasing expressiveness induces an initial bias around *flat output functions*. Notably, this initial solution can be very close to satisfying many physics PDEs, i.e., falling into a local minimum of the PINN loss that only minimizes PDE residuals, while still being far from the true solution that jointly minimizes PDE residuals and the initial and/or boundary conditions. It is difficult for gradient descent optimization to escape from such a local minimum trap, often causing the training to stall. We then prove that the sinusoidal mapping of inputs—in an architecture we label as sf-PINN—is effective to increase input gradient variability, thus avoiding being trapped in such deceptive local minimum. The level of variability can be effectively modulated to match high-frequency patterns in the problem at hand. A key facet of this paper is the comprehensive empirical study that demonstrates the efficacy of learning in sinusoidal spaces with PINNs for a wide range of forward and inverse modelling problems spanning multiple physics domains.

## I. INTRODUCTION

THEORY-GUIDED machine learning has been drawing increasing interest in recent years [1]–[4]. Physics-informed neural networks—PINNs as per Raissi *et al.* [3]—in particular have leveraged the expressiveness of deep neural networks (DNNs) to model the dynamical evolution $\hat{u}(x, t; \boldsymbol{w})$ of physical systems in space $x \in \Omega$ and time $t \in [0, T]$, via the optimization of network parameters $\boldsymbol{w}$. In many of these physical systems, $u$ can be mathematically described by a known prior derived from the underlying physics, such as governing equations in the form of ordinary differential equations (ODEs) and partial differential equations (PDEs). The uniqueness of a PINN lies in incorporating the residual term for such PDEs into the training loss function. This physics-augmented loss thus acts as a penalty to constrain the PINN from violating the PDE, ensuring that its output obeys underlying governing physics.

There has been a recent surge in PINN studies for various science and engineering problems, including in the fields of fluid mechanics [5]–[11], solid mechanics [12]–[14] and optics [15]–[17]. These problems have one common characteristic: ground truth data is difficult and expensive to obtain (often requiring extensive domain expertise), whether it be through computational simulations or real-world experiments. In such settings, the imposition of physics-based constraints can reduce overfitting to sparse data by ensuring the consistency of a PINN's outputs with fundamental physics laws [18]. Given the resultant proficiency in learning from limited (or even *zero*) labelled data, variants of PINNs have begun to attract attention for their utility in the following three areas:

1. PINNs form a new class of mesh-free methods to solve PDEs, i.e., the *forward problems* [19]–[26]. Essentially, the problem is transformed to one of neural network training requiring no labelled data and no explicit discretization. The loss function is simply defined by residual terms from the PDEs and prescribed initial conditions (IC) or boundary conditions (BC).

2. PINNs have been extended to solve *inverse problems*, i.e., inferring *unknown* physics in PDEs from limited data [27]. For example, they have been successfully applied to quantify fluid flows from visualization or sensor data [5]–[7], and in the design of metamaterials [15], [16].

3. Another extension to the forward problem is to construct a *meta-PINN* model $\hat{u}(x, t, \omega; \boldsymbol{w})$, where $\omega$ represents different scenarios of the input condition (initial or boundary condition), geometry, or the differential equation parameters [10], [11], [28]–[31]. The meta-PINNs can be directly learned from the physics law either without any labelled data or when supplemented with sparse data.

Jian Cheng Wong is with the Institute of High Performance Computing (IHPC), Agency for Science, Technology and Research (A*STAR), Singapore, and is also with the School of Computer Science and Engineering, Nanyang Technological University (NTU), Singapore (email: wongj@ihpc.a-star.edu.sg).

Chinchun Ooi is with the Institute of High Performance Computing (IHPC) and Centre for Frontier AI Research (CFAR), Agency for Science, Technology and Research (A*STAR), Singapore (email: ooicc@ihpc.a-star.edu.sg).

Abhishek Gupta is with the Singapore Institute of Manufacturing Technology (SIMTech), Agency for Science, Technology and Research (A*STAR), Singapore (email: abhishek_gupta@simtech.a-star.edu.sg).

Yew-Soon Ong is Chief Artificial Intelligence Scientist with the Agency for Science, Technology and Research (A*STAR), Singapore, and is also with the Data Science and Artificial Intelligence Research Centre, School of Computer Science and Engineering, Nanyang Technological University (NTU), Singapore (email: asysong@ntu.edu.sg).

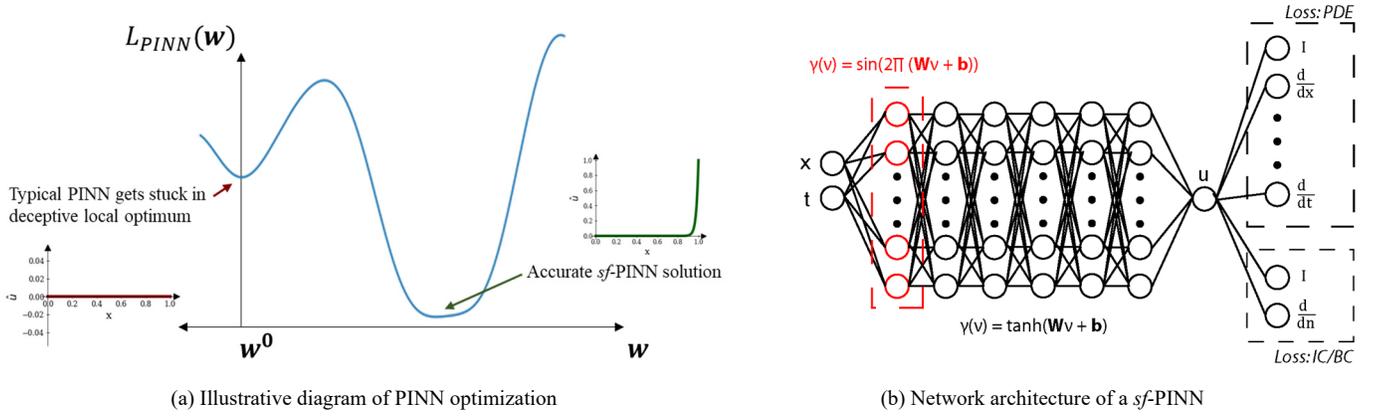

(a) Illustrative diagram of PINN optimization

(b) Network architecture of a *sf*-PINN

Fig. 1. (a) Illustrative diagram showing how a typical PINN model is initialized at a local minimum, and hence can get stuck during training. (b) Network architecture of a *sf*-PINN with sinusoidal feature mapping appended to a typical PINN with *tanh* activation in subsequent hidden layers.

Despite the strong practical motivations in a wide range of applications, it remains difficult to train accurate PINN models. For example, in many works, training iterations for PINNs far exceed those needed for typical DNNs [3], [32], [33] even when using state-of-the-art stochastic gradient descent (SGD) methods. As a result, there has been significant effort of late to improve PINN trainability by balancing the different components in its loss function [34]–[39]. However, this is a highly non-trivial task as the individual components vary greatly in magnitude and convergence rate during training.

Given the above, in this paper, we present a comprehensive investigation of the merits of learning in *sinusoidal space* for accurate PINNs. The paper contains three thematic portions. First, we theoretically examine the behavior of the input gradients of a typical PINN at initialization and relate this to its trainability issue. *Going beyond the vanishing gradient problem of DNNs* [40]*, we uncover a major reason for the difficulty in training PINNs stems from the model's strong initial bias towards flat output functions (with zero input gradients almost everywhere)*. This is particularly undesirable for PINNs since a flat function minimizes the residuals of many commonly encountered physics PDE. As a result, the PINN gets trapped in a deceptive local optimum (that only minimizes PDE residual terms) right from the onset of training, while still being far away from the true solution that must also satisfy initial and/or boundary conditions; see Figure 1a.

Based on this revelation, a new perspective on the utility of learning in sinusoidal spaces is laid out next; in particular, how they extricate PINNs from the local optimum trap. We introduce a class of sinusoidal mappings and theoretically analyze the effectiveness of initializing PINNs with appropriate input gradient distribution. The resultant neural architecture is referred to as *sf*-PINN; see Figure 1b. The gradients' variability boost in *sf*-PINN takes effect across the entire input domain, as opposed to a typical PINN where only input gradients near origin are sensitive to the change in weights distribution. This property critically affects PINN training since the samples are usually uniformly drawn from the input domain and are equally important. It is also worth noting the ubiquity of sinusoidal structures in analytical solutions of dynamical systems such as electromagnetics and advection-diffusion problems, which suggests their natural suitability for learning effective representations for PINNs.

Lastly, our comprehensive experimental study demonstrates that *sf*-PINNs are indeed able to achieve better performance (a few orders of magnitude improvement in accuracy) than standard PINNs across a wide range of forward and inverse modelling problems, spanning multiple physics domains. Interestingly, the results also suggest that *sf*-PINNs are less sensitive to the choice of network architecture and initialization methods, as well as the composition of PINN loss components. This makes finding the right weightage for PINN loss components less critical, offering an alternate path to improving the quality of PINN than those in current literature.

The remainder of the paper is organized as follows. In Section II, we describe training issues faced by a typical PINN model. Section III examines the theoretical behavior of a typical PINN at initialization and its implication during training. Section IV introduces sinusoidal mappings as an approach to mitigate the uncovered theoretical limitations. Section V enumerates the experimental studies conducted to verify and illustrate the advantage of *sf*-PINN across multiple forward and inverse modelling problems. Concluding remarks and direction for future research are then presented in Section VI.

## II. STANDARD PINNs AND THEIR LIMITATIONS

### A. Overview of a typical PINN

We briefly outline the PINN methodology as commonly employed in current literature and software [41]–[43]. A typical PINN considers a multilayer perceptron (MLP) representation, $\hat{u}(x, t; \boldsymbol{w})$, for modelling the dynamical process $u$ of a physical system in space $x \in \Omega$ and time $t \in [0, T]$, with network parameters $\boldsymbol{w}$ to be optimized. The spatial domain usually has 1-, 2- or 3-dimensions in physical problems. Furthermore, $u$ mathematically obeys known priors such as PDEs of the general form [3], [23]:

$$\mathcal{N}_t[u(x, t)] + \mathcal{N}_x[u(x, t)] = 0, \quad x \epsilon \Omega, t \epsilon [0, T], \quad (1a)$$

$$u(x, 0) = u_o(x), \quad x \epsilon \Omega, \quad (1b)$$

$$\mathcal{B}[u(x, t)] = g(x, t), \quad x \epsilon \partial\Omega, t \epsilon [0, T], \quad (1c)$$

where $\mathcal{N}_t[\cdot]$ is the temporal derivative, and $\mathcal{N}_x[\cdot]$ is a general nonlinear differential operator which can include any combination of non-linear terms of spatial derivatives, such as the first and second order derivatives $\frac{\partial u}{\partial x}$ and $\frac{\partial^2 u}{\partial x^2}$ respectively. The IC at $t=0$ is defined by $u_o(x)$, and the boundary operator $\mathcal{B}[\cdot]$ enforces the desired condition $g(x,t)$ at the domain boundary $\partial\Omega$. This $\mathcal{B}[\cdot]$ can be an identity operator (Dirichlet BC) or a differential operator (Neumann BC).

As an example, consider the wave equation—describing phenomena such as electromagnetic propagation—which in 1D reads $\frac{\partial^2 u}{\partial t^2} = c^2\, \frac{\partial^2 u}{\partial x^2}$, where $\mathcal{N}_t[u] = \frac{\partial^2 u}{\partial t^2}$ and $\mathcal{N}_x[u] = -c^2\, \frac{\partial^2 u}{\partial x^2}$. It contains a coefficient $c$ representing the velocity of the wave. The PDE in isolation has trivial solutions $u(x,t) = \mu$, where $\mu$ is any constant, among many other solutions. A nontrivial solution must therefore be obtained by satisfying the PDE (1a) in combination with the IC (1b) and/or BC (1c) as prescribed by the problem of interest.

### 1) PINN training loss

The loss function of a PINN is defined as [3], [23]:

$$\mathcal{L}_{PINN} = \mathcal{L}_{Data} + \lambda_{PDE}\mathcal{L}_{PDE} + \lambda_{IC}\mathcal{L}_{IC} + \lambda_{BC}\mathcal{L}_{BC}, \tag{2a}$$

which includes the data loss component (if data is available),

$$\mathcal{L}_{Data} = \frac{1}{n}\sum_{i=1}^{n}(u_i - \hat{u}_i)^2, \tag{2b}$$

and the physics loss components,

$$\mathcal{L}_{PDE} = \|\hat{u}_t(\cdot\,;\boldsymbol{w}) + \mathcal{N}_x[\hat{u}(\cdot\,;\boldsymbol{w})]\|^2_{\Omega\times[0,T]}, \tag{2c}$$

$$\mathcal{L}_{IC} = \|\hat{u}(\cdot\,,0\,;\boldsymbol{w}) - u_0\|^2_{\Omega}, \tag{2d}$$

$$\mathcal{L}_{BC} = \|\mathcal{B}[\hat{u}(\cdot\,;\boldsymbol{w})] - g(\cdot)\|^2_{\partial\Omega\times[0,T]}. \tag{2e}$$

The relative weights, $\lambda$s, in (2) control the trade-off between different components in the loss function and need to be scaled depending on the problem at hand. The computation of the PINN loss involves matching the PINN output $\hat{u}$ against target $u$ over $n$ labelled samples (2b), substitution of the output $\hat{u}$ into the PDEs for evaluating the residuals (2c) over the input domain $\Omega\times[0,T]$, as well as matching the output $\hat{u}$ against IC at $t=0$ (2d), and BC over the domain boundary $\partial\Omega$ (2e). In solving forward problems, the data loss component $\mathcal{L}_{Data}$ (2b) may be omitted. The physics loss component (2c-e) is defined over a continuous domain, but for practical reasons, we compute the residuals over a finite set of $m$ collocation points $D = \{(x_i, t_i)\}_{i=1}^{m}$. These points are sampled, for example, using a uniform grid or randomized Latin hypercube sampling. Differential operators, such as $\frac{\partial \hat{u}}{\partial t}$, $\frac{\partial \hat{u}}{\partial x}$, $\frac{\partial^2 \hat{u}}{\partial x^2}$, are required for the evaluation of the PDEs residuals. For a PINN that is higher order differentiable, the computation of differential operators can then be conveniently obtained via automatic differentiation [44].

### B. Empirical observation of training issue

PINNs are found to be more challenging to train than usual DNNs, which has so far hindered its full potential in broad applications. For demonstration, we consider solving a 1D

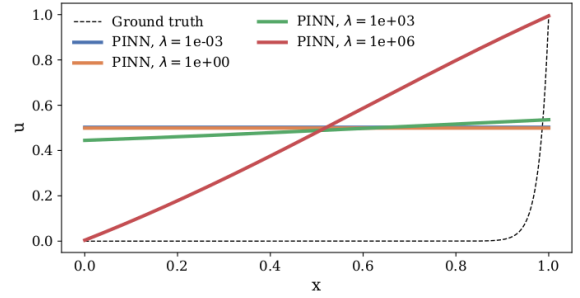

Fig. 2. Training results of a typical PINN with different $\lambda$ values used in PINN loss (4), for the 1D convection-diffusion equation. The PINN outputs are given in colored lines. The dashed line indicates the ground truth solution.

convection-diffusion equation,

$$v\,\frac{\partial u}{\partial x} = k\,\frac{\partial^2 u}{\partial x^2}, \quad x\epsilon[0,L], \tag{3}$$

with the PDE parameters $v=50$, $k=1$. Here, we set the BC $u(x=0)=0$; $u(x=1)=1$ ($L=1$). We adopt a MLP (4 hidden layers) with *tanh* activation and initialize the network weights by *Xavier* method [45]—a typical PINN implementation widely used in the literature [3], [41]–[43]. We attempt to solve the problem (3) by optimizing network weights $\boldsymbol{w}$ w.r.t. the following PINN loss,

$$\underset{\boldsymbol{w}}{\text{minimize}}\; \lambda^{-1}\mathcal{L}_{PDE} + \mathcal{L}_{BC}\; (= \mathcal{L}_{PINN}), \tag{4}$$

using a popular SGD variant – the ADAM optimizer [46].

The training results when using different $\lambda$ values are shown in Figure 2, suggesting a typical PINN can face difficulties in solving this (seemingly) simple 1D problem. Tuning $\lambda$ makes the PINN training prioritize different loss terms representing the PDE and BC, thus its solution jumps from a flat line (minimizes $\mathcal{L}_{PDE}$) to a slope (minimizes $\mathcal{L}_{BC}$). However, they are all far away from the true solution that minimizes both $\mathcal{L}_{PDE}$ and $\mathcal{L}_{BC}$ even upon careful selection of $\lambda$.

## III. ANALYSIS OF THE STANDARD PINN MODEL

### A. Behavior at initialization of a typical PINN

A MLP architecture with *tanh* activation is widely adopted for PINNs because it is higher order differentiable, hence the differential operators can be conveniently evaluated during training. The *Xavier* initialization is a popular and proven method for such architecture in many DNN applications, therefore it is also commonly adopted as *default* in many PINNs. We demonstrated in Section II-B with a simple example that this typical choice for PINNs is likely to yield poor results.

To gain insight into factors causing the training failure, we analyze the initial input gradient distribution of a typical PINN with $L$ hidden layers ($L$ usually ranges between 2 and 10) and $n$ neurons per layer. It will be convenient to denote the initial outputs at hidden layer $l$ as,

$$\boldsymbol{u}_l = \boldsymbol{W}_l \boldsymbol{x}_l \quad \textit{(before activation)}, \tag{5a}$$

$$\boldsymbol{x}_{l+1} = f(\boldsymbol{u}_l) \quad \textit{(after activation)}. \tag{5b}$$

Here, $\boldsymbol{x}_l \epsilon \mathbb{R}^{n\times 1}$ is a real vector consisting of $n$ inputs (denoted

$x_l$'s) to the hidden layer, and $\boldsymbol{W}_l \epsilon \mathbb{R}^{n \times n}$ is a real matrix consisting of $n \times n$ weights (denoted $w_l$'s). The $w_l$'s are assumed to be independently and identically distributed (i.i.d.). $f$ is the activation function applied to each element $u_l$ in $\boldsymbol{u}_l \epsilon \mathbb{R}^{n \times 1}$, and $\boldsymbol{x}_{l+1} \epsilon \mathbb{R}^{n \times 1}$ consists of $n$ inputs to the next hidden layer. Without loss of generality, we denote the single (1D) input and output of a $L$-hidden layers PINN as $x_1 \equiv x$ and $\hat{u} \equiv u_L = \boldsymbol{W}_L \boldsymbol{x}_L$. Note that the full set of network parameters $\boldsymbol{w}$ thus represents $\{\boldsymbol{W}_1, \boldsymbol{W}_2, ..., \boldsymbol{W}_L\}$. The input gradient $\frac{\partial \hat{u}}{\partial x}$ plays an important role in PINN training because the loss is formulated based on such differential operators (see Figure 1b). We can derive the input gradient for arbitrary input $x$ using the chain rule,

$$\frac{\partial \hat{u}}{\partial x} = \sum_{j=1}^{n} w_{L,j} f'(u_{L-1,j}) \frac{\partial u_{L-1,j}}{\partial x}, \tag{6a}$$

$$\frac{\partial u_{l,j}}{\partial x} = \sum_{i=1}^{n} w_{l,ij} f'(u_{l-1,i}) \frac{\partial u_{l-1,i}}{\partial x}, \text{ for } 1 < l < L, \tag{6b}$$

$$\frac{\partial u_{1,i}}{\partial x} = w_{1,i}, \tag{6c}$$

where $x_{l,j} = f(u_{l,j})$ is the $j$-th input of the $l$-th hidden layer (also $j$-th output of the $(l-1)$-th hidden layer). We also assume that $w_l$'s are independent of $\frac{\partial u_{l-1}}{\partial x}$, where $\frac{\partial u_{l-1}}{\partial x}$'s are identically distributed. The initial behavior and immediate consequence of such a PINN is presented in the following Proposition 1.

*Proposition 1. Let $\hat{u}(x; \boldsymbol{w})$ be a PINN with $L$ fully connected layers, $n$ neurons per layer, activation function $f$=tanh, and parameters $\boldsymbol{w}$. Let all the dense layers be initialized by the Xavier method for network weights, i.e., $w_l$'s are i.i.d. $\mathcal{N}(0, \frac{2}{fan_{in} + fan_{out}})$, where $fan_{in}$ and $fan_{out}$ are the number of inputs and outputs for the dense layer. Then, as $n \to \infty$, $\hat{u}(x; \boldsymbol{w})$ trivially satisfies arbitrary differential equations of the form $F\left(\frac{\partial u}{\partial x}, \frac{\partial^2 u}{\partial x^2}, ..., \frac{\partial^k u}{\partial x^k}\right) = 0$ where $F(\varphi_1, \varphi_2, \varphi_3, ...) = \sum_i a_i \varphi_i + \sum_{i \leq j} b_{ij} \varphi_i \varphi_j + \sum_{i \leq j \leq k} c_{ijk} \varphi_i \varphi_j \varphi_k + \cdots$, with probability 1 at initialization.*

*Proof:*

The mean of the input gradient at initialization is given by,

$$\mathrm{E}\left[\frac{\partial \hat{u}}{\partial x}\right] = n \mathrm{E}\left[w_L f'(u_{L-1}) \frac{\partial u_{L-1}}{\partial x}\right] = n \mathrm{E}[w_L] \mathrm{E}\left[f'(u_{L-1}) \frac{\partial u_{L-1}}{\partial x}\right] = 0, (7)$$

since $w_l \sim \mathcal{N}(0, \frac{2}{n+1})$. The result also holds for other $\frac{\partial u_l}{\partial x}$, i.e., $\mathrm{E}\left[\frac{\partial u_l}{\partial x}\right] = 0$ for $1 \leq l \leq L$. The variance of the input gradient at initialization is then given by,

$$\mathrm{var}\left(\frac{\partial \hat{u}}{\partial x}\right) = n \, \mathrm{var}\left(w_L f'(u_{L-1}) \frac{\partial u_{L-1}}{\partial x}\right) =$$
$$n \, \mathrm{var}(w_L) \mathrm{E}\left[\left(f'(u_{L-1}) \frac{\partial u_{L-1}}{\partial x}\right)^2\right] \leq n \, \mathrm{var}(w_L) \, \mathrm{var}\left(\frac{\partial u_{L-1}}{\partial x}\right) =$$
$$\frac{2n}{n+1} \mathrm{var}\left(\frac{\partial u_{L-1}}{\partial x}\right), \tag{8}$$

since $0 < f'(u_l) \leq 1$ for $1 \leq l < L$ with $f' = \mathrm{sech}^2$. Backward passing this result through previous hidden layers $1 < l < L$,

$$\mathrm{var}\left(\frac{\partial u_l}{\partial x}\right) \leq n \, \mathrm{var}(w_l) \, \mathrm{var}\left(\frac{\partial u_{l-1}}{\partial x}\right) = \mathrm{var}\left(\frac{\partial u_{l-1}}{\partial x}\right), \tag{9}$$

under *Xavier* initialization, i.e., $w_l \sim \mathcal{N}(0, 1/n)$. Note that at the first hidden layer, $\mathrm{var}\left(\frac{\partial u_1}{\partial x}\right) = \frac{2}{n+1}$ since $w_l \sim \mathcal{N}(0, \frac{2}{n+1})$. Hence, we can derive the variance of input gradient,

$$\mathrm{var}\left(\frac{\partial \hat{u}}{\partial x}\right) \leq \frac{2n}{n+1} \mathrm{var}\left(\frac{\partial u_{L-1}}{\partial x}\right) \leq \frac{2n}{n+1} \mathrm{var}\left(\frac{\partial u_{L-2}}{\partial x}\right) \leq \cdots \leq$$
$$\frac{2n}{n+1} \mathrm{var}\left(\frac{\partial u_2}{\partial x}\right) \leq \frac{2n}{n+1} \mathrm{var}\left(\frac{\partial u_1}{\partial x}\right) = \frac{2n}{n+1} \frac{2}{n+1}, \tag{10}$$

which is inversely proportional to $n$, with $\lim_{n \to \infty} \mathrm{var}\left(\frac{\partial \hat{u}}{\partial x}\right) = 0$. The above results imply $\frac{\partial \hat{u}}{\partial x} \xrightarrow{P} \mathrm{E}\left[\frac{\partial \hat{u}}{\partial x}\right] = 0$ when $n \to \infty$, $\forall x$.

A PINN with *zero* input gradient implies a constant output, i.e., $\hat{u}(x; \boldsymbol{w}) = \mu$. This further implies all higher order derivatives $\frac{\partial^2 \hat{u}}{\partial x^2}, \frac{\partial^3 \hat{u}}{\partial x^3}, ..., \frac{\partial^k \hat{u}}{\partial x^k}$ must all equal to 0. It follows that $F\left(\frac{\partial \hat{u}}{\partial x}, \frac{\partial^2 \hat{u}}{\partial x^2}, ..., \frac{\partial^k \hat{u}}{\partial x^k}\right) = 0$. Hence, differential equations of the form stated in the Proposition are trivially satisfied at initialization. ∎

Although we only present the derivations for a single input, single output PINN in Proposition 1, the extension to multiple inputs, e.g., $\boldsymbol{x} = [x, y, t]^\mathrm{T}$, and outputs is straightforward. Note that differential equations of the form stated in Proposition 1 encapsulate many classical physics equations encountered in the real world such as the heat equation, wave equation, Laplace equation, to name just a few [47]. The *zero* input gradient condition also trivially satisfies important nonlinear PDEs [48], such as the Navier-Stokes (N-S) equation [49] without external source, which govern complex fluid physics.

### B. Implication to PINN training

Proposition 1 indicates that a very expressive PINN, i.e., one with width $n \to \infty$, can have its initial input gradient collapse to *zero*. This makes the PINN trivially minimize the residual term of many practical PDEs at initialization. The result also suggests that a typical PINN with moderate depth and width can have a near *zero* input gradient, ref. to (10), in which case the output is not far from satisfying the PDE loss (2c), when initialized. This initial bias can make the PINN training using gradient descent methods challenging in at least one fundamental way.

The successful training of PINN is essentially predicated on *matching both the initial/boundary conditions and the PDEs in the entire input domain*, ref. to (2). The PINN described in Proposition 1 falls right into the local minimum trap of the PINN loss by only minimizing the PDE loss term; see Figure 1a for illustration. However, the true solution that jointly minimizes the PDE loss and BC loss can be very far away from the initialization in weights space. This is particularly the case when the true solution consists of high frequencies and/or steep gradients. In such cases, it will be *difficult for gradient descent optimization to escape from the deceptive local minimum trap* given that the PINN already has a near *zero* initial input gradient distribution, such that sensible solutions are not accessible near the initialization during the training process.

Another issue stems from the known saturation of the *tanh* activation function as the network weights grow large during training, which also causes gradients to fall to *zero* and hence impede further network training (by gradient-based optimization). Such loss of input gradient information at initialization and during training is inappropriate for PINN problems, especially those with higher frequencies and steep gradients. For example, it causes a poorly optimized PINN solution to the simple PDE problem presented in Section II-B.

The training challenges faced by PINNs is somewhat different from those encountered by usual data-fit DNN models. For example, the initial bias towards near *zero* input gradient has not been an issue for data-fit DNN models, and the learning bias towards lower frequency outputs has been accepted as a natural regularizer for large data-fit DNN models with potential benefits to their generalization performance [50], [51]. In the context of PINN, the generalization performance is usually de-prioritized over obtaining the global minimum to the PINN loss. These biases with potential benefit to DNN models actually hinder the efficacy of finding a good candidate solution to simultaneously minimize all loss terms during PINN training.

### C. Other activation and initialization methods for PINNs

The analysis from previous sections indicates that a near *zero* initial input gradient distribution should be avoided in PINNs. We therefore consider alternate activation and initialization methods with the aim of elevating the input gradients. Notably, the results stated in Proposition 1 also extend to other smooth activation function with the property $f'(y) \leq 1$, such as $f = sin$ and $f = sigmoid$. For $f = sigmoid$, $f'(u_i) \leq 1/4$, thus the initial input gradient distribution converges to *zero* even quicker than *tanh* or *sin*.

As compared to *Xavier*, another popular initialization—*He* method [52]— initializes network weights with higher variance, i.e., $w_l \sim \mathcal{N}(0, \frac{2}{fan_{in}})$ for layer $l$ with $fan_{in}$ input nodes, which effectively prevents diminishing $\text{var}\left(\frac{\partial \hat{u}}{\partial x_i}\right)$ during the backward pass. A critical difference between *He* and *Xavier* methods is in how they initialize the first hidden layer weights $w_1$'s. Taking a single input PINN as example, *He* gives $w_1 \sim \mathcal{N}(0, 2)$ as opposed to $w_1 \sim \mathcal{N}(0, \frac{2}{1+fan_{out}})$ given by *Xavier*, and the increase in input gradient distribution can positively impact PINN training. Besides, *sin* activation always retains a higher proportion of gradients even when the training happens in the large weight regime, ref. to Figure 3, thereby circumventing the diminishing gradient issue.

Although there are methods such as *He* initialization that can provide a larger initial input gradient, they do not necessarily offer sufficient gradient variability at the beginning of training for achieving an accurate PINN model. Hence, we further explore the strategy of controlling the input gradients of a PINN, such as using larger and problem-specific initial weight distributions. We first present the consequence of initializing a typical PINN with large weight distribution at its input layer, as the following Proposition 2.

*Proposition 2. Let $\hat{u}(x; \boldsymbol{w})$ be the typical PINN described in Proposition 1, <u>except at input layer</u>, the network weights, i.e., the $w_1$'s, are initialized as i.i.d. $\mathcal{N}(0, \sigma^2)$. Then, $\hat{u}(x; \boldsymbol{w})$ almost surely has zero input gradient when $\sigma \to \infty, x \neq 0$ or $|x| \to \infty, \sigma > 0$, at initialization.*

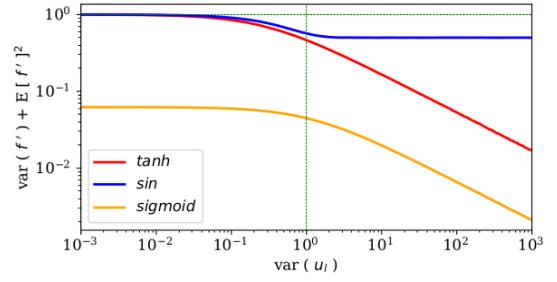

Fig. 3. Simulation results showing the relationship between $\text{var}(u_i)$ and $\text{var}(f'(u_i)) + \text{E}[f'(u_i)]^2$ during backward pass for $f = tanh, sin, sigmoid$, under the scenario $u_i \sim N(0, \text{var}(u_i))$. The results indicate that $\text{var}(f'(u_i)) + \text{E}[f'(u_i)]^2 \approx 1$ when $\text{var}(u_i)$ is small, i.e., below 5e-2. When $\text{var}(u_i)$ is large, $f = sin$ retains the highest proportion of $\text{var}(f'(u_i)) + \text{E}[f'(u_i)]^2 \approx 0.5$ during the backward pass.

*Proof:*

The variance of the input gradient at initialization can be derived as,

$$\text{var}\left(\frac{\partial \hat{u}}{\partial x}\right) \leq \frac{2n}{n+1} \text{var}\left(\frac{\partial u_2}{\partial x}\right) = \frac{2n}{n+1} n \, \text{var}(w_2 f'(u_1) w_1) = \frac{2n}{n+1} n \, \text{var}(w_2) \, \text{E}\left[\left(w_1 f'(u_1)\right)^2\right] = \frac{2n}{n+1} \text{E}[w_1^2 f'(w_1 x)^2], \quad (11)$$

with $f' = sech^2$. We want to show that $\text{E}[w_1^2 f'(w_1 x)^2] \to 0$ as $\text{var}(w_1) \equiv \sigma \to \infty, x \neq 0$ or $|x| \to \infty, \sigma > 0$.

It can be shown by the law of the unconscious statistician,

$$\text{E}[w_1^2 f'(w_1 x)^2] = \int_{-\infty}^{\infty} w_1^2 f'(w_1 x)^2 \frac{1}{\sigma \sqrt{2\pi}} e^{-\frac{1}{2}\left(\frac{w_1}{\sigma}\right)^2} \, dw_1 = 2 \int_0^{\infty} w_1^2 f'(w_1 x)^2 \frac{1}{\sigma \sqrt{2\pi}} e^{-\frac{1}{2}\left(\frac{w_1}{\sigma}\right)^2} \, dw_1, \quad (12)$$

by symmetry because the integrand is an even function. We denote $S = \int_0^{\infty} \left(w_1 f'(w_1 x)\right)^2 \frac{1}{\sigma \sqrt{2\pi}} e^{-\frac{1}{2}\left(\frac{w_1}{\sigma}\right)^2} \, dw_1$ and show that,

$$S \leq \frac{1}{\sigma \sqrt{2\pi}} \int_0^{\infty} \left(w_1 f'(w_1 x)\right)^2 \, dw_1, \text{ because } e^{-\frac{1}{2}\left(\frac{w_1}{\sigma}\right)^2} \leq 1, \quad (13a)$$

$$S \leq \frac{1}{\sigma \sqrt{2\pi}} \int_0^{\infty} \left(4 w_1 e^{-2 w_1 |x|}\right)^2 \, dw_1, \quad (13b)$$

$$S \leq \frac{1}{\sigma \sqrt{2\pi}} \frac{1}{2x^2 |x|}, |x| > 0, \sigma > 0. \quad (13c)$$

The step from (13a) to (13b) comes from the fact $w_1 f'(w_1 x) = w_1 \left(\frac{2}{e^{w_1 x} + e^{-w_1 x}}\right)^2 \leq w_1 \left(\frac{2}{e^{w_1 |x|}}\right)^2 = 4 w_1 e^{-2 w_1 |x|}, w_1 \geq 0$.

Based on (13c), $\lim_{\sigma \to \infty} S \leq 0, x \neq 0$ or $\lim_{|x| \to \infty} S \leq 0, \sigma > 0$. We know from the positivity of integrand that, $S \geq 0$. Hence, we get $\lim_{\sigma \to \infty} S = 0, x \neq 0$ or $\lim_{|x| \to \infty} S = 0, \sigma > 0$. Substituting this result back to (11), we show that the input gradient $\frac{\partial \hat{u}}{\partial x}$ converges to its mean $\text{E}\left[\frac{\partial \hat{u}}{\partial x}\right] = 0$ for finite $n$ with probability 1 as $\sigma \to \infty, x \neq 0$ or $|x| \to \infty, \sigma > 0$, as stated in Proposition 2. ∎

The relationship between $\text{var}(w_1)$ and $\text{E}[w_1^2 f'(w_1 x)^2]$ in (11) is visualized in Figure 4a. Proposition 2 indicates that initializing the weights at the input layer of a typical PINN with a large variance can be ineffective because it inherently reduces the input gradient. Moreover, the results suggest that changing the weight distribution for such PINNs disproportionately affects the input gradient for different $x$ values, i.e., the input gradient converges to *zero* more quickly as $x$ moves away from 0; note the upper bound in the variance of $\frac{\partial \hat{u}}{\partial x}$ as suggested by (13c). This also explains why the input gradients near the origin are most sensitive to the weights change during PINN training, while becoming insensitive for $|x| \gg 0$. This property affects PINN training more critically as compared to usual data-fit DNN models because PINN training samples are usually uniformly drawn from the input domain and are equally important everywhere due to its physical correspondence to the spatial-temporal extents of the problem, as opposed to common deep learning tasks where samples are normally distributed, i.e., samples are dense near the origin of the input.

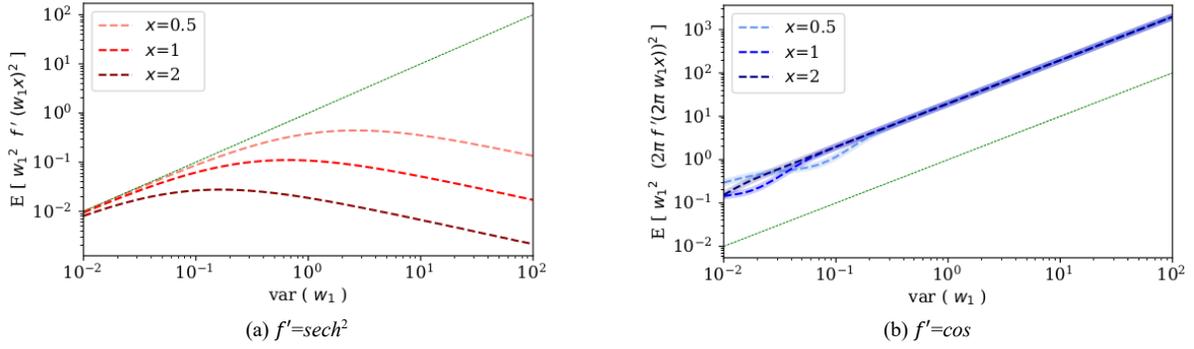

(a) $f'=sech^2$

(b) $f'=cos$

Fig. 4. Simulation results showing the relationship between $\mathrm{var}(w_1)$ and (a) $\mathrm{E}[w_1^2 f'(w_1 x)^2]$ for $f$=$tanh$ and (b) $\mathrm{E}\left[w_1^2\big(2\pi f'(2\pi w_1 x)\big)^2\right]$ for *sinusoidal feature mapping* described in (14). For $f$=$tanh$, the relationship is approximately linear when $\sigma$ or $|x|$ is small, while approaching *zero* as they get large, thus causing the input gradient $\frac{\partial \hat{u}}{\partial x} \to 0$ as described in Proposition 2. For *sinusoidal feature mapping*, the relationship is approximately linear. The shaded lines in (b) are obtained from the analytical relationship (16) given by Proposition 3.

## IV. PINN TRAINING IN SINUSOIDAL SPACES

Hence, we postulated that an appropriate initial input gradient distribution is important to successful PINN training. Therefore, we need a convenient and effective method to initialize PINN with appropriate input gradient distribution. We show that this can be achieved through sinusoidal feature mapping of the PINN inputs $\boldsymbol{x} = [x,t]^\mathsf{T}$, and hence learning in sinusoidal spaces:

$$\gamma(\boldsymbol{x}) = \sin\big(2\pi(\boldsymbol{W_1}\boldsymbol{x}+\mathbf{b_1})\big), \qquad (14)$$

where $\boldsymbol{W_1} \epsilon \mathbb{R}^{n_1 \times d}$ is a real matrix related to the frequency of sinusoidal features and $\mathbf{b_1} \epsilon \mathbb{R}^{n_1 \times 1}$ is a real vector related to phase lag. Here, $n_1$ is a user-defined number of sinusoidal features, and $d$ is the number of inputs. We can incorporate the sinusoidal mapping $\gamma(\boldsymbol{x})$ into the input layer of a PINN; see Figure 1b. This is essentially replacing the input layer of a typical PINN described in Proposition 2 with $\gamma(\boldsymbol{x})$. The weight and bias parameters of the input layer now become the mapping parameters and they can be optimized during PINN training. This architecture is referred to hereafter as *sf-PINN*.

We initialize the input layer weights $w_1$'s in $\boldsymbol{W_1}$ by sampling from a $\mathcal{N}(0,\sigma^2)$, and the bias as *zero*. The idea is to use $\sigma^2 \equiv \mathrm{var}(w_1)$ in the input layer (sinusoidal features mapping), as a way to scale the initial input gradient distribution. The importance of using *sin* in (14) is highlighted in Proposition 3.

*Proposition 3.* Let $\hat{u}(x;\boldsymbol{w})$ be the *sf-PINN*, i.e., the <u>input layer</u> of the typical PINN described in Proposition 1 is <u>replaced</u> by sinusoidal feature mapping $\gamma(x)$ described in (14) with $n$ features/nodes, and the network weights $w_1$'s initialized as i.i.d. $\mathcal{N}(0,\sigma^2)$. Then, $\hat{u}(x;\boldsymbol{w})$ has an initial input gradient variance with upper bound $\frac{2n}{n+1}2\pi^2\sigma^2[1+e^{-8\pi^2\sigma^2 x^2}(1-16\pi^2\sigma^2 x^2)]$, $0 < \sigma < \infty$.

*Proof:*

The variance of input gradient for *sf-PINN* at initialization can be derived as,

$$\mathrm{var}\left(\frac{\partial \hat{u}}{\partial x}\right) \leq \frac{2n}{n+1}\mathrm{var}\left(\frac{\partial u_2}{\partial x}\right) = \frac{2n}{n+1}n\,\mathrm{var}(w_2 w_1 2\pi f'(2\pi\,w_1 x)) = \frac{2n}{n+1}\mathrm{E}\left[w_1^2\big(2\pi f'(2\pi w_1 x)\big)^2\right], \qquad (15)$$

with $f'$=$cos$. We can further derive,

$$\mathrm{E}\left[w_1^2\big(2\pi f'(2\pi\,w_1 x)\big)^2\right] =$$
$$\int_{-\infty}^{\infty} w_1^2\big(2\pi\,cos(2\pi\,w_1 x)\big)^2 \frac{1}{\sigma\sqrt{2\pi}}e^{-\frac{1}{2}\left(\frac{w_1}{\sigma}\right)^2}\,dw_1 = 2\pi^2\sigma^2\big[1+ e^{-8\pi^2\sigma^2 x^2}(1-16\pi^2\sigma^2 x^2)\big],\ 0 < \sigma < \infty, \qquad (16)$$

Substituting results (16) back to (15), we get,

$$\mathrm{var}\left(\frac{\partial \hat{u}}{\partial x}\right) \leq \frac{2n}{n+1}2\pi^2\sigma^2\big[1+e^{-8\pi^2\sigma^2 x^2}(1-16\pi^2\sigma^2 x^2)\big], \qquad (17)$$

$0 < \sigma < \infty$, as stated in Proposition 3. ∎

The relationship between $\mathrm{var}(w_1)$ and $\mathrm{E}\left[w_1^2\big(2\pi f'(2\pi w_1 x)\big)^2\right]$ in (15) is visualized in Figure 4b. According to Proposition 3, the *sf-PINN* has a maximum input gradient variance $\mathrm{var}\left(\frac{\partial \hat{u}}{\partial x}\right) = \frac{2n}{n+1}4\pi^2\sigma^2$ at $x=0$, and a near constant upper bound $\frac{2n}{n+1}2\pi^2\sigma^2$ when $|x| \gg 0$ or $\sigma \gg 0$. Equality in (17) occurs for a 1 hidden layer *sf-PINN*. Hence, the increase in $\sigma^2$, i.e., $\mathrm{var}(w_1)$, of the input layer weights in a *sf-PINN*, can effectively elevate the gradient everywhere in the input domain. With this property, we can prescribe better initialization for *sf-PINNs* with sufficient input gradient variability to escape the local minimum trap during early stages of training. An appropriate initial gradient distribution also makes the training easier since the weights distribution does not need to be abruptly changed. Such *sf*-PINNs also become less sensitive to the activation and initialization schemes used in subsequent hidden layers. We will empirically demonstrate these benefits in Section V.

There is another benefit of using the sinusoidal features mapping described in (14). Sinusoidal patterns occur naturally in numerous physical processes, for example, in analytical descriptions of wave propagation and solutions to the heat equation. For such problems, the sinusoidal features provide *sf*-PINNs an informed guess to the target solution, and potentially transforms the loss landscape into an easier one to optimize. *sf*-PINN may more easily find the desired sinusoidal patterns (input gradients profile) that match the inherent physical characteristics of the target solution, thus achieving significant speedup in training over PINNs. It's worth noting that sinusoidal features do not necessarily constrain the output to be periodic within the spatial-temporal domain of interest.

## V. EXPERIMENT RESULTS

### A. *sf*-PINN with different activations and initialization schemes

The effectiveness of *sf*-PINN for circumventing local minima in PINN loss and quicker convergence to an accurate solution is demonstrated with 2 example PDE problems: *1) 1D steady-state convection-diffusion equation*, and *2) 2D steady-state N-S equations (lid-driven cavity)*. In each problem, we compare the solution MSE of the *sf*-PINNs to *standard*-PINNs with different activation (*tanh*, *sin*, *sigmoid*) and initialization (*Xavier*, *He*) methods. The model architecture and training settings are reported in TABLE I.

### 1) 1D steady-state convection diffusion equation

We solve the problem described in (3). The solution to this problem is relatively simple, with a steep gradient near one end ($x_1 = 1$) that needs to be obeyed by PINN. The key factor to obtaining an accurate solution is that the PINN training needs

TABLE I: PINN MODEL AND OPTIMIZATION CONFIGURATIONS USED IN EXPERIMENTAL STUDY 4.4.1.

| Test problem | 1) 1D steady-state convection-diffusion eqn. | 2) 2D steady-state N-S eqns. (lid-driven cavity) |
|---|---|---|
| PINN architecture | $(x)$–**32**–10–10–10–$(\hat{u})$ | $(x, y)$–**64**–20–20–20–[20–20–20–$(\hat{u})$), 20–20–20–$(\hat{v})$, 20–20–20–$(\hat{p})$] |
| Training sample | 5000 | 52x52 |
| Batch size | 499 + 1 | 475 + 25 |
| Training iteration | 50,000 | 200,000 |
| Initial learning rate | 5e-3 | 1e-3 |
| $\lambda, \sigma$ (*sf*-PINN) | $\lambda$=500, $\sigma$=0.5 | $\lambda$=1, $\sigma$=1 |

1. For the PINN architecture, the numbers in between input and output represent the number of nodes in each hidden layer. For example, $(x)$–**32**–10–10–10–$(\hat{u})$ indicates a single input $x$, followed by 4 hidden layers with 32, 10, 10 and 10 nodes in each layer, and a single output $\hat{u}$.
2. *sf*-PINN replaces the <u>input layer</u> with the sinusoidal features mapping (14) and initialize its weights by sampling from a $\mathcal{N}(0, \sigma^2)$.
3. Batch size: points sampled for 1 evaluation of $\mathcal{L}_{PINN} = \lambda^{-1}\mathcal{L}_{PDE} + \mathcal{L}_{BC}$.
4. A training iteration: 1 evaluation of $\mathcal{L}_{PINN}$ for backpropagating the weight gradients. We update PINN weights in every 100 iterations. We reduce the learning rate on plateauing, until a min. learning rate of 1e-6 is reached.

to escape from the local minimum at initialization and find a right path to descend.

### 2) 2D steady-state N-S equations (lid-driven cavity)

The lid-driven cavity problem has been widely chosen as a benchmark case for numerical and PINN methods, due to the complex physics encapsulated within. This problem is a unit square cavity with a lid velocity $u_{lid} = 1$ for the top wall, while other walls are non-slip. The governing equations are the 2D steady-state incompressible N-S equations:

$$\frac{\partial u}{\partial x} + \frac{\partial v}{\partial y} = 0, \tag{18a}$$

$$u\frac{\partial u}{\partial x} + v\frac{\partial u}{\partial y} = \frac{1}{Re}\left(\frac{\partial^2 u}{\partial x^2} + \frac{\partial^2 u}{\partial y^2}\right) - \frac{\partial p}{\partial x}, \tag{18b}$$

$$u\frac{\partial v}{\partial x} + v\frac{\partial v}{\partial y} = \frac{1}{Re}\left(\frac{\partial^2 u}{\partial x^2} + \frac{\partial^2 u}{\partial y^2}\right) - \frac{\partial p}{\partial y}. \tag{18c}$$

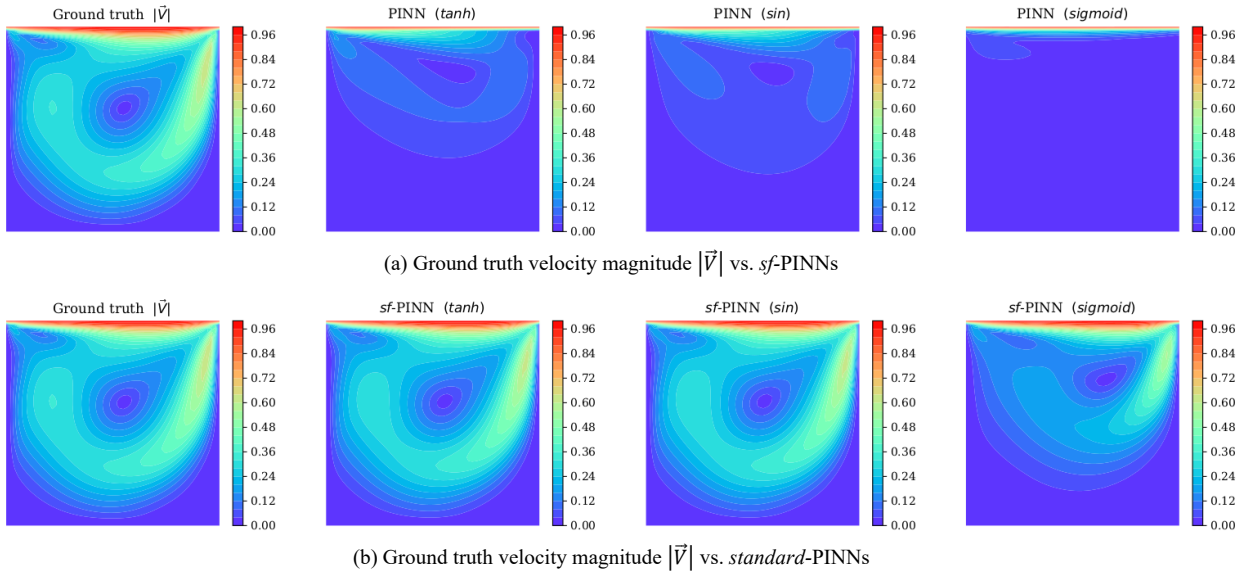

(a) Ground truth velocity magnitude $|\vec{V}|$ vs. *sf*-PINNs

(b) Ground truth velocity magnitude $|\vec{V}|$ vs. *standard*-PINNs

Fig. 5.  Contour plot of the ground truth velocity magnitude $|\vec{V}|$ and solutions obtained from (a) *sf*-PINNs and (b) *standard*-PINNs with different activation (*tanh*, *sin*, *sigmoid*) and initialized by Xavier method, for the lid driven cavity problem at $Re = 400$. PINN results are averaged from 20 independent runs.

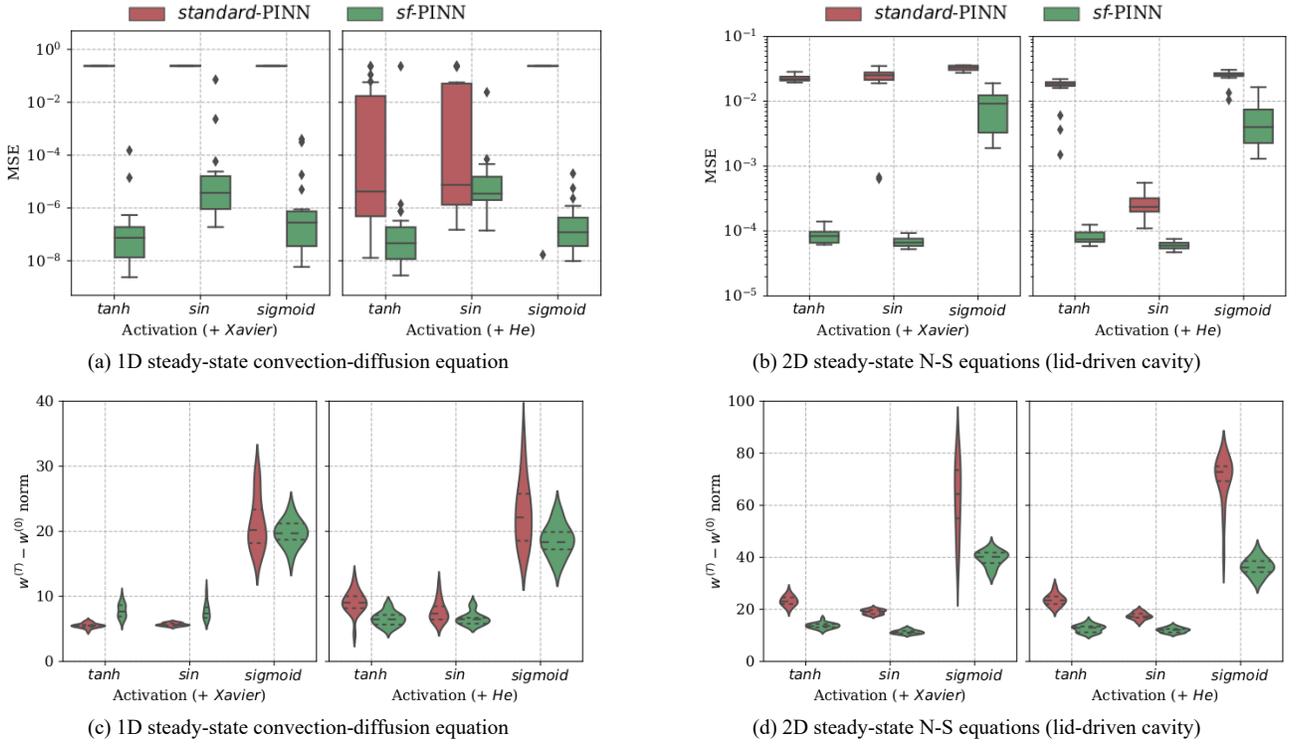

Fig. 6. Distribution of solution MSE between PINNs and ground truth for the problems (a) 1D steady-state convection-diffusion equation and (b) 2D steady-state N-S equations (lid-driven cavity), and (c-d) their weights' displacement norm between initialization and final PINNs after training. In (b), we computed the MSE for $u$- and $v$-velocity components and take the average. Results are aggregated from 20 independent runs.

In the above equations, the primitive variables are velocity $(u, v)$ and pressure $p$. $Re$ is Reynolds number which represents the ratio of inertial forces to viscous forces. We solve for the lid-driven cavity problem at $Re$=400; there will be a primary vortex at the center region and 2 eddies at the bottom-corners. To compute the MSE for the PINN solutions, the ground truth is obtained by an in-house numerical solver based on the improved divergence-free condition compensated (IDFC) method [53].

*3) Discussion of results*

The distribution of solution MSE based on 20 independent runs for different PINNs are summarized in Figure 6a-b. For the *convection-diffusion* problem, all *Xavier* initialized *standard*-PINNs fail to obtain a good solution regardless of their activation. Switching to *He* initialization improves the *standard*-PINNs' training results, except for those with *sigmoid* activation. The main difference between *Xavier* and *He* methods were discussed in Section III-C, and this change improves the *standard*-PINNs' training result. *sf*-PINNs display significantly improved training relative to corresponding *standard*-PINNs across different activation and initialization schemes. In this example, *sf*-PINN with *tanh* activation in subsequent hidden layers is most accurate.

For the *lid-driven cavity* problem, only *He* initialized *standard*-PINNs with *sin* activation converge to a plausible solution, whereas other *standard*-PINNs get stuck in bad solution. *sf*-PINNs significantly improve the training results relative to their corresponding *standard*-PINNs across different activation and initialization schemes. In this example, *sf*-PINN with *sin* activation in subsequent hidden layers produces a slightly more accurate solution than *tanh*. Figure 5 visually compares the velocity magnitude $|\vec{V}| = \sqrt{u^2 + v^2}$ contour computed from solutions solved by selected *standard*-PINNs and *sf*-PINNs, against the simulated ground truth.

Overall, the sinusoidal features mapping significantly improves PINNs' performance. The results evince the importance of having appropriate initial gradient distribution. Thus, *sf*-PINNs are made easier to train and converge to a more accurate solution while requiring less adaptation to their weights from initialization as compared to *standard*-PINNs, as evidenced in Figure 6c-d. In contrast, the poor MSE observed for certain *standard*-PINNs despite a relatively large adaptation in weights indicate the stiff training issue associated with large weight distributions, especially in more complex PDE problems such as the lid-driven cavity example. Consistent with the explanations in Section III-C, the *sigmoid* is commonly outperformed by the other activations.

### B. sf-PINN with different sinusoidal features mapping methods

The sinusoidal features mapping method described in Section IV is general and it encapsulates other novel variants. Below, we introduce notable examples from the literature that are related to our *sf*-PINN framework:

1. Sinusoidal Representation Network (SIRENs) [54]. PINNs with *sin* activation—SIRENs as per Sitzmann *et al.* [54]—can be seen as a specific implementation of *sf*-PINNs with the *sin* activation adopted in all subsequent hidden layers and initialized by *He* method. The introduction of *sin* activation in DNNs and association to Fourier neural

TABLE II: Comparison between standard-PINN and sf-PINN Variants Used in Experimental Study 4.4.2.

| | Input layer | $(\mathbf{W_1}, \mathbf{b_1})$ trainable? | Subsequent hidden layers Activation | Initialization |
|---|---|---|---|---|
| *standard*-PINN | $\gamma(\boldsymbol{x}) = \tanh(\mathbf{W_1}\boldsymbol{x} + \mathbf{b_1})$, $\mathbf{W_1} \epsilon \mathbb{R}^{n_1 \times d}, \boldsymbol{x} \epsilon \mathbb{R}^{d \times 1}, \mathbf{b_1} \epsilon \mathbb{R}^{n_1 \times 1}$ | Yes | *tanh* | *Xavier* |
| *sf*-PINN | $\gamma(\boldsymbol{x}) = \sin(2\pi(\mathbf{W_1}\boldsymbol{x} + \mathbf{b_1}))$, $\mathbf{W_1} \epsilon \mathbb{R}^{n_1 \times d}, \boldsymbol{x} \epsilon \mathbb{R}^{d \times 1}, \mathbf{b_1} \epsilon \mathbb{R}^{n_1 \times 1}$ | Yes | *tanh* | *Xavier* |
| SIREN | | Yes | *sin* | *He* |
| *ff*-PINN | $\gamma(\boldsymbol{x}) = \left[ \sin(2\pi(\mathbf{W_1}\boldsymbol{x} + \mathbf{b_1})), \ \cos(2\pi(\mathbf{W_1}\boldsymbol{x} + \mathbf{b_1})) \right]^{\mathrm{T}}$, | Yes | *tanh* | *Xavier* |
| *rf*-PINN | $\mathbf{W_1} \epsilon \mathbb{R}^{\frac{n_1}{2} \times d}, \boldsymbol{x} \epsilon \mathbb{R}^{d \times 1}, \mathbf{b_1} \epsilon \mathbb{R}^{\frac{n_1}{2} \times 1}$ | No | *tanh* | *Xavier* |

TABLE III: PINN Model and optimization Configurations Used in Experimental Study 4.4.2.

| Problem | 1) 1D transient wave eqn. | 2) 2D transient N-S eqns. (Taylor–Green vortex) | 3) 1D transient KdV eqn. | 4) 2D Helmholtz eqn. |
|---|---|---|---|---|
| PINN architecture | $(x,t)$–**64**–50–50–50–$(\hat{u})$ | $(x,y,t)$–**64**–50–50–50–[50–50–50–$(\hat{u})$, 50–50–50–$(\hat{v})$, 50–50–50–$(\hat{p})$] | $(x,t)$–**64**–50–50–50–$(\hat{u})$ | $(x,y)$–**64**–20–20–20–$(\hat{u})$ |
| Training sample | 256×256 | 101x101x51 | 257×251 | 256×256 |
| Batch size | 450 + 40 + 10 | 450 + 40 + 10 | 480 + 10 + 10 | 450 + 0 + 50 |
| Training iteration | 200,000 | 100,000 | 100,000 | 100,000 |

1. Batch size: points sampled for 1 evaluation of $\mathcal{L}_{PINN} = \lambda^{-1}\mathcal{L}_{PDE} + \mathcal{L}_{IC} + \mathcal{L}_{BC}$.

2. A training iteration: 1 evaluation of $\mathcal{L}_{PINN}$ for backpropagating the weight gradients. We update PINN weights every 100 iterations. The initial learning rate for PINN optimization is set at 5e-3. We reduce the learning rate on plateauing, until a min. learning rate of 1e-6 is reached.

3 For the inverse problems described in Section V-C, the points sampled for 1 evaluation of $\mathcal{L}_{PINN} = \mathcal{L}_{Data} + \lambda^{-1}\mathcal{L}_{PDE}$ (batch size) is set as 50 + 450.

network date back to the 1980s [55]. However, they typically do not offer superior performance relative to other activations in common machine learning tasks and have therefore become more uncommon in recent literature. Lately, several works in literature reported a superior performance with sinusoidal activations in PINNs [5], [25], [26], without systematically exploring the benefit of sinusoidal activation or elucidating why it performs better than other activation functions on PINN problems.

2. Fourier features PINNs (*ff*-PINNs) [56]. The *ff*-PINNs replace (31) with pairs of *sin* and *cos* (a constant phase shift of *sin*) mapping of the inputs, i.e., $\gamma(\boldsymbol{x}) = \left[ \sin(2\pi(\mathbf{W_1}\boldsymbol{x} + \mathbf{b_1})), \ \cos(2\pi(\mathbf{W_1}\boldsymbol{x} + \mathbf{b_1})) \right]^{\mathrm{T}}$. Both *sin* and *cos* components share the same mapping parameters $(\mathbf{W_1}, \mathbf{b_1})$. As suggested in recent studies [56], [57], Fourier mapping of inputs has a significant effect on the NTK's eigenvalue spectrum and convergence properties of the corresponding DNN/PINN.

3. Random Fourier features PINNs (*rf*-PINNs) [57]. The *rf*-PINNs have the exact same setting as the *ff*-PINNs, except that the randomly initialized Fourier features (i.e., *sin* and *cos* mappings) are not subsequently fine-tuned, i.e., $(\mathbf{W_1}, \mathbf{b_1})$ are frozen during training.

According to our analysis in previous sections, we hypothesize that each of these variants of the sinusoidal mappings are functionally similar, and they will have similar efficacy when applied to PINNs. In the following, we benchmark the performance of *sf*-PINNs using the sinusoidal features mapping described in (14) and above variants (*ff*-PINNs, *rf*-PINNs, SIRENs) against a typical PINN (*standard*-PINNs), see TABLE II, as a mesh free method to solve a wide range of PDE problems. The test problems include *1) 1D transient wave equation, 2) 2D transient N-S equations (Taylor–Green vortex), 3) 1D transient KdV equation, and 4)*

*2D Helmholtz equation.* For each test problem, we train different *sf*-PINN models under consistent settings, as reported in TABLE III. Then, we compute their MSE against the ground truth solution over a dense set of test points which are sampled from the problem domain.

The use of sinusoidal features mapping incurs an additional hyperparameter $\sigma$ to tune, which we also refer to as the *bandwidth* parameter; it has an immediate impact on the initial gradient distribution at the *sf*-PINN's input and is associated with the frequency range of the sinusoidal features. By adjusting $\sigma$, we can modulate the initial gradient distribution to better match the frequency characteristic of the problem at hand. In practice, if the output frequency is unknown, the $\sigma$ needs to be selected by a grid search. To demonstrate the effect of $\sigma$ in *sf*-PINNs and its variants vs. *standard*-PINNs, we run a series of experiments with different bandwidth parameter $\sigma \epsilon [1e\text{-}1, 1e1]$. Moreover, the convergence of PINNs is highly sensitive to the choice of relative weights $\lambda$s in PINN loss (2). We further compare the performance of these PINN models for a consistent $\sigma = \sigma_{fixed}$ and different relative weights $\lambda \epsilon [1, 1e6]$ in the loss function, i.e., $\mathcal{L} = \lambda^{-1}\mathcal{L}_{PDE} + \mathcal{L}_{IC} + \mathcal{L}_{BC}$. We uniformly sample 25 different $\sigma$ and 25 different $\lambda$ in log space.

### 1) 1D transient wave equation

The wave equation describes the wave phenomena arising in the physical world such as water waves, sound waves, and seismic waves. We consider the wave equation in 1D:

$$\frac{\partial^2 u}{\partial t^2} = c^2 \frac{\partial^2 u}{\partial x^2}, \tag{19a}$$

where $c$ is a coefficient representing the velocity of the wave. We apply PINNs to solve equation (19a) with $c$=2 and the following IC and BC,

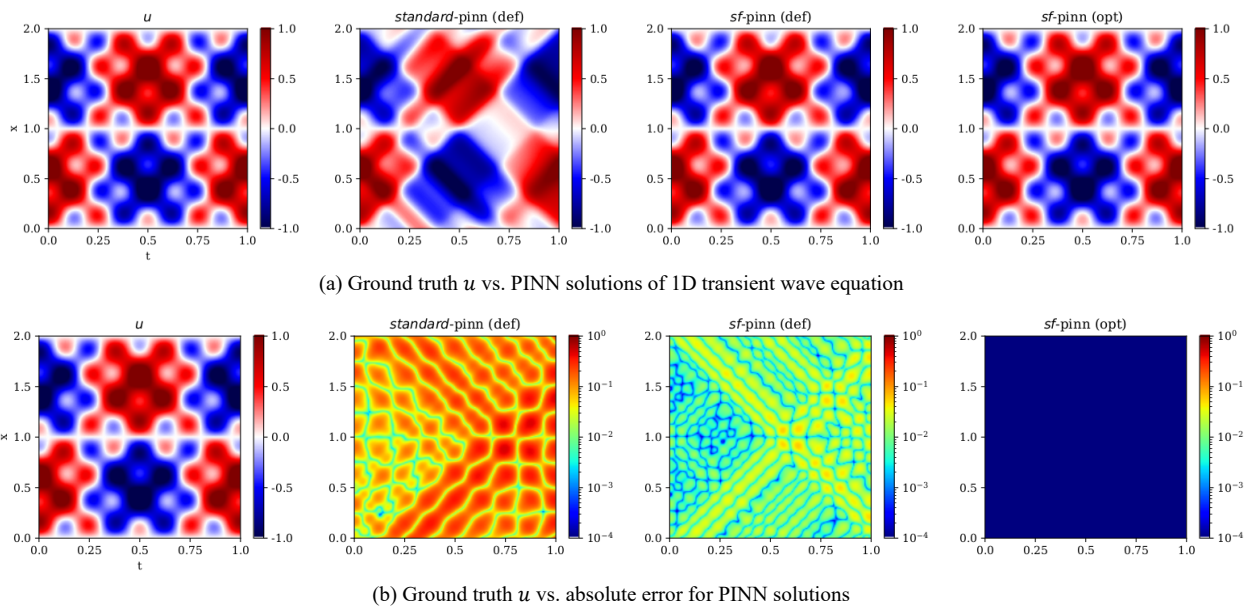

(a) Ground truth $u$ vs. PINN solutions of 1D transient wave equation

(b) Ground truth $u$ vs. absolute error for PINN solutions

Fig. 7. (a) Ground truth solution of 1D transient wave equation and the solutions solved by *standard*-PINN and *sf*-PINN using default setting $\sigma = 1, \lambda = 1$ (def), and *sf*-PINN using optimized setting $\sigma = 2.5, \lambda = 180$ (opt). (b) Absolute error between ground truth and PINN solutions. Results are aggregated from 5 independent runs.

$$u(x,0) = \sin(\pi x) + 0.5 \sin(4\pi x), \tag{19b}$$
$$u_t(x,0) = 0, \tag{19c}$$
$$u(x_L,t) = u(x_U,t) = 0, \tag{19d}$$

solved within domain $x\epsilon[x_L, x_U], t\epsilon[0,1]$ where $x_L = 0, x_U = 2$. The above scenario leads to an oscillatory solution,

$$u(x,t) = \sin(\pi x)\cos(c\pi t) + 0.5\sin(4\pi x)\cos(4c\pi t). \tag{20}$$

The comparison between solution obtained from a *standard*-PINN and a *sf*-PINN is visualized in Figure 7. A *standard*-PINN using default $\sigma$=1, $\lambda$=1 gives poor results. The contour plot shows a significant deviation from the ground truth (MSE=2e-

2). The multi-frequency waves are not resolved correctly. On the other hand, a fairly accurate solution (MSE=2.6e-4) can be obtained by using a *sf*-PINN of the same setting. The accuracy of the *sf*-PINN solution greatly improves (MSE<1e-10) when an optimized $\sigma$ and $\lambda$ configuration is found. Importantly, the solution from *standard*-PINN models remain worse even after a similar search for an optimal $\lambda$.

### 2) 2D transient N-S equations (Taylor–Green vortex)

The N-S equations are a set of PDEs describing the motion of viscous fluid—which describes the physics of many phenomena of scientific and engineering interest. The 2D transient incompressible N-S equations are given by

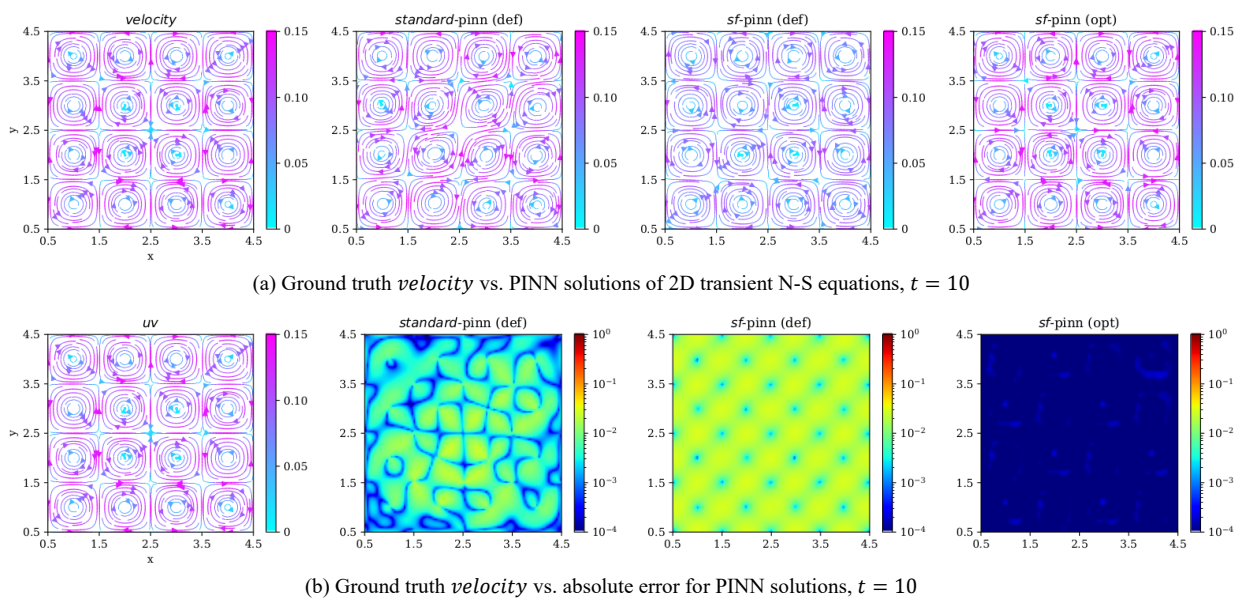

(a) Ground truth *velocity* vs. PINN solutions of 2D transient N-S equations, $t = 10$

(b) Ground truth *velocity* vs. absolute error for PINN solutions, $t = 10$

Fig. 8. (a) Ground truth solution of 2D transient N-S equations (Taylor–Green vortex) at $t = 10$ and the solutions solved by *standard*-PINN and *sf*-PINN using default setting $\sigma = 1, \lambda = 1$ (def), and *sf*-PINN using optimized setting $\sigma = 0.68, \lambda = 1$ (opt). The vortices are represented by streamlines, and the velocity magnitude is indicated by the color. (b) Absolute error between ground truth and PINN solutions. Results are aggregated from 5 independent runs.

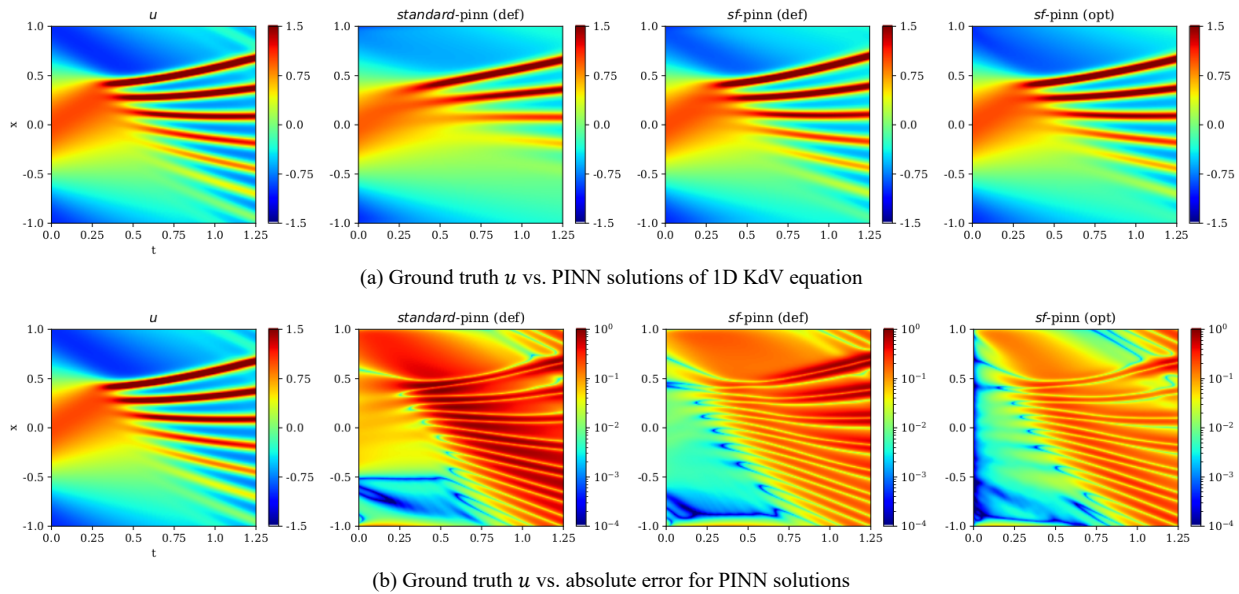

(a) Ground truth $u$ vs. PINN solutions of 1D KdV equation

(b) Ground truth $u$ vs. absolute error for PINN solutions

Fig. 9. (a) Ground truth solution of 1D transient KdV equation and the solutions solved by *standard*-PINN and *sf*-PINN using default setting $\sigma = 1, \lambda = 1$ (def), and *sf*-PINN using optimized setting $\sigma = 1, \lambda = 180$ (opt). (b) Absolute error between ground truth and PINN solutions. Results are aggregated from 5 independent runs.

$$\frac{\partial u}{\partial x} + \frac{\partial v}{\partial y} = 0, \tag{21a}$$

$$\frac{\partial u}{\partial t} + u\frac{\partial u}{\partial x} + v\frac{\partial u}{\partial y} = \frac{1}{Re}\left(\frac{\partial^2 u}{\partial x^2} + \frac{\partial^2 v}{\partial y^2}\right) - \frac{\partial p}{\partial x}, \tag{21b}$$

$$\frac{\partial u}{\partial t} + u\frac{\partial v}{\partial x} + v\frac{\partial v}{\partial y} = \frac{1}{Re}\left(\frac{\partial^2 v}{\partial x^2} + \frac{\partial^2 v}{\partial y^2}\right) - \frac{\partial p}{\partial y}, \tag{21c}$$

where $t$ is time, $u, v$ are the flow velocity, $p$ is the pressure, and $Re$ is Reynolds number which represents the ratio of inertial forces to viscous forces. In particular, the Taylor–Green vortex serves as a popular benchmark for testing and validation of numerical methods for solving incompressible N-S equations. The problem is an unsteady flow of decaying vortices which has the exact closed form solution:

$$u(x, y, t) = -\cos(\pi x)\sin(\pi y)\exp(-2\pi^2 \nu t), \tag{22a}$$

$$v(x, y, t) = \sin(\pi x)\cos(\pi y)\exp(-2\pi^2 \nu t), \tag{22b}$$

$$p(x, y, t) = -\frac{\rho}{4}\left[\cos(2\pi x) + \cos(2\pi y)\right]\exp(-4\pi^2 \nu t). \tag{22c}$$

The problem is solved within the spatial-temporal domain $x\epsilon[0.5, 4.5]$, $y\epsilon[0.5, 4.5]$, $t\epsilon[0, 10]$ with $Re=100$, by using $u(x, y, 0), v(x, y, 0), p(x, y, 0)$ as the initial condition. At the 4 spatial domain boundaries, a Dirichlet BC for $u, v$ and a Neumann BC for $p$ are specified.

In this test problem, the MSE is computed for the Euclidean norm of 2D velocity vector $|\vec{V}| = \sqrt{u^2 + v^2}$ between ground truth and PINN solution. Although the *standard*-PINN using default $\sigma=1, \lambda=1$ could achieve a solution (MSE~1e-5), the flow visualization (Figure 8) of velocity streamlines at the final time step reveals that the vortices in the center have been diverted and their shapes are distorted. On the other hand, the vortices are well maintained in the solution given by a *sf*-PINN using the same default setting. With an appropriately chosen $\sigma$, the MSE of a *sf*-PINN solution drops significantly to below 1e-9. The pattern and magnitude of the vortices are almost indistinguishable from the ground truth. Similar to the wave equation, the solution from *standard*-PINN models remain worse even after a similar search for an optimal configuration.

### 3) 1D transient KdV equation

The KdV equation is used in physics and engineering to model the weakly nonlinear long waves (e.g., on shallow water surfaces). We consider the KdV equation in the form:

$$\frac{\partial u}{\partial t} + u\frac{\partial u}{\partial x} + \nu\frac{\partial^3 u}{\partial x^3} = 0, \tag{23}$$

where the coefficient $\nu$ is set to be 0.0005. We apply PINNs to solve equation (23) with the IC $u(x, 0) = \cos(\pi x)$ and with the periodic BC, within the domain $x\epsilon[-1, 1], t\epsilon[0, 1.25]$. We compared the PINN solutions against simulation results obtained by an in-house numerical solver as ground truth. As shown in Figure 9, the solution manifests a waveform evolving into multiple waves over time. This is a difficult problem for PINNs to solve. The solution obtained from *standard*-PINN with default $\sigma=1, \lambda=1$ is increasingly erroneous at later $t$. The *sf*-PINN with the same default setting offers improvement to the overall solution. The primary flow features are resolved better, and more fine-scale waves can be seen. Nevertheless, the solution is still not very accurate (MSE~1e-2) even after optimizing for $\sigma$ and $\lambda$, highlighting the limitation of present PINN approach.

### 4) 2D Helmholtz equation

The Helmholtz equation is an important PDE in many physics and engineering systems such as optics, acoustics, electrostatics, and quantum mechanics. We first consider a Helmholtz equation of the following form:

$$\frac{\partial^2 u}{\partial x^2} + \frac{\partial^2 u}{\partial y^2} + u = q(x, y), \tag{24a}$$

$$q(x, y) = (1 - \pi^2 - (6\pi)^2)\sin(\pi x)\sin(6\pi y). \tag{24b}$$

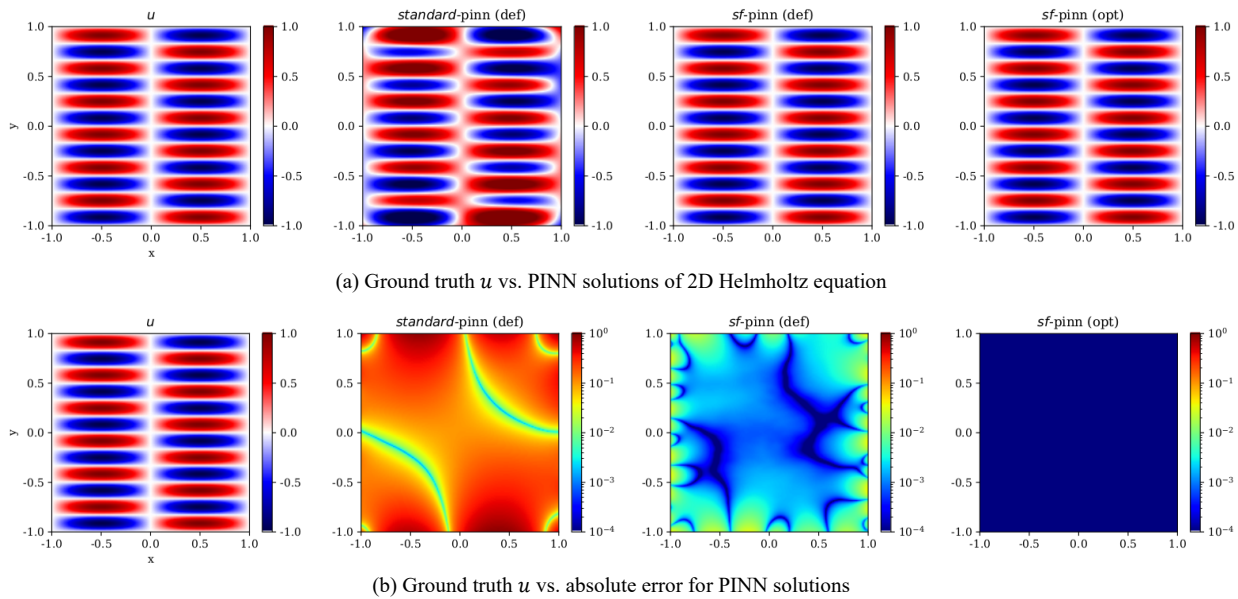

(a) Ground truth $u$ vs. PINN solutions of 2D Helmholtz equation

(b) Ground truth $u$ vs. absolute error for PINN solutions

Fig. 10. (a) Ground truth solution of 2D Helmholtz equation and the solutions solved by *standard*-PINN and *sf*-PINN using default setting $\sigma = 1, \lambda = 1$ (def), and *sf*-PINN using optimized setting $\sigma = 2.5, \lambda = 1000$ (opt). (b) Absolute error between ground truth and PINN solutions. Results are aggregated from 5 independent runs.

where $q(x, y)$ is the source term. The equation (24) is solved by PINNs within the input domain $x \epsilon[-1, 1]$, $y \epsilon[-1, 1]$, with a Dirichlet BC at the 4 domain boundaries. Its solution is the multiplication of 2 sinusoids with different frequency:

$$u(x, y) = \sin(\pi x) \sin(6\pi y). \tag{25}$$

Figure 10 compares the ground truth solution with the solutions obtained from *standard*-PINN and *sf*-PINNs. A *standard*-PINN with default $\sigma$=1, $\lambda$=1 fails to properly solve the problem. Obvious artifacts can be observed in the solution. A *sf*-PINN with the same default setting greatly improves the MSE by 3 orders of magnitude, i.e., from 5.6e-2 to 2.4e-5. A significantly better *sf*-PINN model can be obtained by increasing $\sigma$ to accommodate higher frequencies, for example setting $\sigma$=2.5 leads to a very accurate solution (MSE=1.1e-10). This optimal *sf*-PINN solution is indistinguishable from the ground truth, with absolute error less than 1e-4 almost everywhere within the domain.

We further show that it is possible to infer a suitable bandwidth $\sigma$ from the fundamental frequencies of the problem (if known) without running a grid search. We note that this test problem has a frequency of $\sin(6\pi y)$. We then compute the probability for a sinusoidal feature to be initialized close to that frequency. The probability trends are visualized in Figure 11. Given that there are 64 sinusoidal features in the *sf*-PINN model, the probability of having at least 1 of the sinusoidal features initialized with a frequency close to $\sin(6\pi y)$ increases sharply when $\sigma$ is chosen between 1 and 6. The *sf*-PINNs with $\sigma$ chosen within this region are more easily trained, hence leading to significantly more accurate solutions. On the other hand, if the $\sigma$ is chosen to be too big, the probability to sample a sinusoidal feature near the $\sin(6\pi y)$ frequency starts decreasing. Then, the *sf*-PINN training may not be as effective.

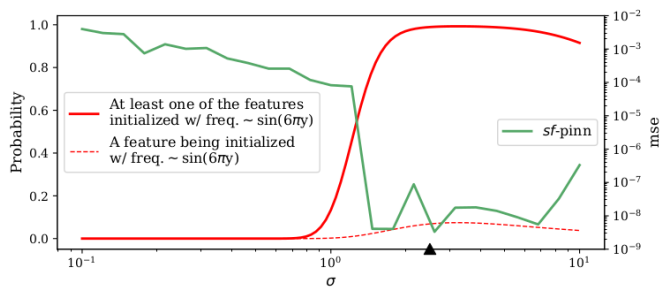

Fig. 11. The probability of having a sinusoidal feature (dashed red line) and at least one of the 64 sinusoidal features (solid red line) initialized with a frequency $\sim \sin(6\pi y)$ given $\sigma \in [1e-1, 10]$. The MSE of sf-PINN against ground truth versus $\sigma$ is also overlaid in the same plot.

*5) Discussion of results across different $\sigma$ and $\lambda$*

The overall results of MSE achieved by different PINNs at various $\sigma$ and $\lambda$ values are compared in Figure 12. The empirical results reveal that the *sf*-PINN and its *ff*-PINN and SIREN variants give very similar performance. These results again evince that the sinusoidal mapping of inputs is an effective method to adjust the initial gradient distribution in PINN, whereas both *tanh* and *sin* (i.e., a SIREN activation function in the subsequent hidden layers) have their own merits. In addition, the full Fourier features mapping used by *ff*-PINN does not outperform *sin* only mapping. We also conclude that there is a strong motivation to optimize the sinusoidal features mapping over the use of random features, as evidenced by the fact that the MSE obtained from *rf*-PINN is consistently worse than other *sf*-PINN variants across almost all test problems.

As expected, the performance of *sf*-PINN and its variants are $\sigma$ and $\lambda$ dependent. Within an appropriate range of $\sigma$, *sf*-PINNs greatly improve in their solution MSE by 3-4 orders of magnitude across the test problems. Empirically, we found that $\sigma$=1 is a practical default, and that the search range can be restricted between 1e-1 and 1e1 for most problems we have encountered. It may also be possible to infer potential $\sigma$ values based on fundamental frequencies of the physical system. On the other hand, the change in $\sigma$ does not significantly affect the performance of *standard*-PINN, since the $\sigma$ parameter does not significantly improve the magnitude of initial gradient distribution. *sf*-PINN remains competitive as compared to *standard*-PINN for most of the $\sigma$ range spanning [1e-1, 1e1]. In addition, the empirical results demonstrate that the superior performance of *sf*-PINN models is consistent across a wide range of $\lambda$, further illustrating robustness to the choice of $\lambda$ in practical implementation. Interestingly, across our range of examples spanning different physics, the *standard*-PINN never out-performs the *sf*-PINN and its variants.

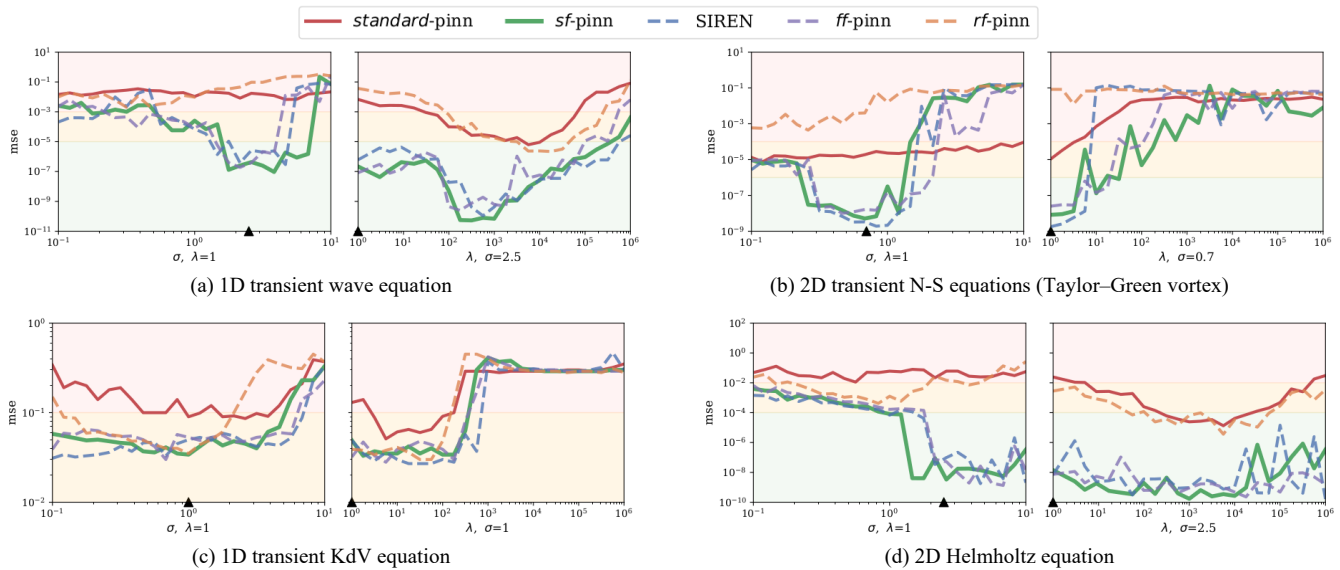

Fig. 12. The best MSE achieved by different PINNs over 5 independent runs, at $\sigma \in [1\text{e-}1, 1\text{e}1]$, $\lambda = \lambda_{fix}$ (left) and $\lambda \in [1, 1\text{e}6]$, $\sigma = \sigma_{fix}$ (right), for different test problems (a-d). The triangle symbol in $\sigma$ plot indicates the $\sigma$ value to be fixed for $\lambda$ experiments, and vice versa. The background color provides a rough guide to the quality of PINNs, where those in the red region present a clear differentiation to the ground truth solution upon visual inspection. A fair solution can be found in yellow region, where some minor artifacts are still noticeable. The accurate solutions in green region are almost indistinguishable from the ground truth.

### C. *sf*-PINN for inverse problem

To investigate the effectiveness of *sf*-PINNs for inverse problem, we carried out experiments on *1) 1D transient wave equation* and *2) 2D transient N-S equations (Taylor–Green vortex)*, as per the forward problems described above. We consider the scenario where some physical property of the governing equations is unknown, i.e., the wave velocity $c$ (ground truth=2) in wave equation (19), and the Reynolds number $Re$ (ground truth=100) in the N-S equations (21). Given the observation data, the task is to infer the unknown parameter. The exact initial and boundary conditions are also unknown, besides what is observed from the data. Our PINN loss for inverse problem takes the form $\mathcal{L} = \mathcal{L}_{Data} + \lambda^{-1} \mathcal{L}_{PDE}$. It is important to note that the collocation points for computing the $\mathcal{L}_{PDE}$ can be arbitrarily sampled within the problem domain, independent of the observed data. The PINN model and training settings are kept identical to the forward problem

We further consider a scenario whereby the observation data is dense, and a scenario whereby the observation data is sparse. In each of the scenarios, we run a series of experiments with different bandwidth parameter $\sigma \in [1\text{e-}1, 10]$ and compare the MSE of the approximated solution given by *sf*-PINNs, SIRENs and *standard*-PINNs. The results are displayed in Figure 13.

The empirical results reveal that both *sf*-PINN and its SIREN variant give very similar performance.

#### 1) 1D transient wave equation

*sf*-PINNs outperform *standard*-PINN in both dense (#observation=256×256) and sparse (#observation=200, with greater concentration at later time) data scenarios. They are particularly beneficial in sparse data scenario when the lack of data increases the difficulty of PINN training—hindering a *standard*-PINN from obtaining an accurate solution. Out of 5 independent runs from a *standard*-PINN ($\sigma=1$), the mean estimated value for the unknown $c$ in the wave equation is 1.999 ($sd=1.7\text{e-}3$). The solution and its absolute error (MSE=6e-4) against the ground truth are plotted in Figure 13c. On the other hand, a very accurate solution (MSE=3.2e-8) can be obtained by using a *sf*-PINN ($\sigma=2.6$). The model also accurately infers the unknown $c=2$ ($sd=7.3\text{e-}6$). The results suggest that the *sf*-PINN can more accurately infer the dynamical behavior backward in time without knowing the complete physics based on sparsely available historical data and the currently observed information, in addition to more accurately inferring the unknown physical parameter.

#### 2) 2D transient N-S equations (Taylor–Green vortex)

Both *sf*-PINN and its SIREN variant achieve similar accuracy for both dense (#observation=101×101×51) and sparse (#observation=600) data scenarios. Figure 13d visualizes the velocity streamlines at the final timestamp, and the absolute error of the solutions obtained by a *standard*-PINN ($\sigma=1$) and a *sf*-PINN ($\sigma=0.83$). The results are inferred from a set of sparse samples which are randomly distributed over the 3D spatial-temporal domain. Their MSEs against the ground truth are 1.4e-5 and 6.9e-9, respectively. Both *standard*-PINN and *sf*-PINN can accurately infer the unknown $\nu = Re^{-1}$ in the N-S equations, with estimates of 9.97e-3 ($sd=3.1\text{e-}5$) and 9.998e-3 ($sd=2.4\text{e-}6$), respectively.

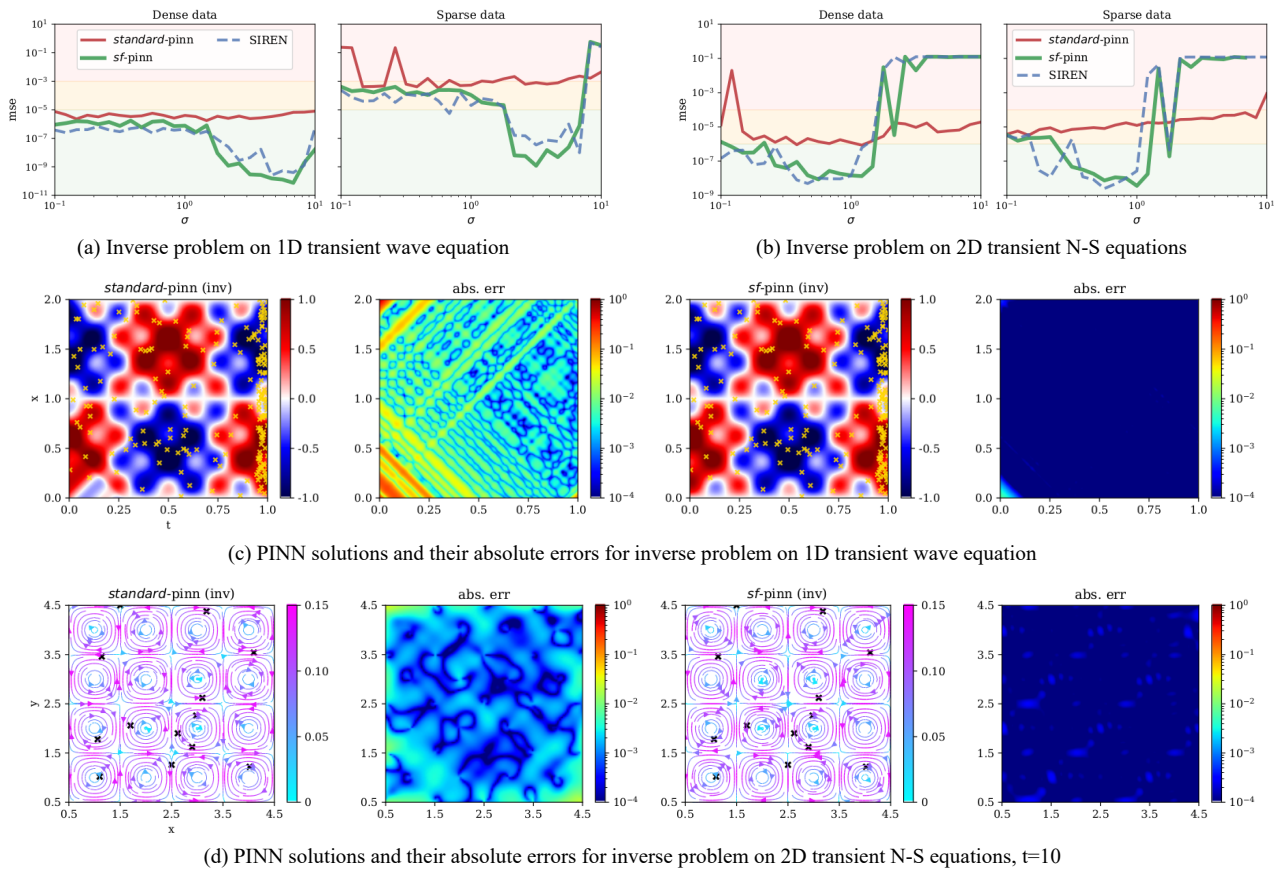

(a) Inverse problem on 1D transient wave equation

(b) Inverse problem on 2D transient N-S equations

(c) PINN solutions and their absolute errors for inverse problem on 1D transient wave equation

(d) PINN solutions and their absolute errors for inverse problem on 2D transient N-S equations, t=10

Fig. 13. The best MSE achieved by different PINNs at $\sigma \in$ [1e-1, 10], for inverse problem on (a) 1D transient wave equation and (b) 2D transient N-S equations, for both scenarios where the data are densely (left) or sparsely (right) observed. The $\lambda$=180 for wave equation and $\lambda$=1 for N-S equations are chosen based on prior experiments for the forward problem. (c) The solutions to wave equation inverse problem solved by a *standard*-PINN ($\sigma$=1, $\lambda$=180) and a *sf*-PINN ($\sigma$=2.6, $\lambda$=180) given sparse data (highlighted in gold-x), and their absolute errors against the ground truth. (d) The solutions to N-S equations inverse problem solved by a *standard*-PINN ($\sigma$=1, $\lambda$=1) and a *sf*-PINN ($\sigma$=0.83, $\lambda$=1) given sparse data (highlighted in black-x), and their absolute errors against the ground truth. Results are aggregated from 5 independent runs.

## VI. CONCLUSION

We analyzed and provided novel perspectives on why a typical PINN frequently exhibits training difficulty, even on a relatively straightforward 1D model problem. We hypothesized this issue originates from the limited initial variability in input gradients of a typical PINN, resulting in a tendency to be trapped in local optimum (that minimizes PDE residual alone) right at the onset of PINN training, while still being far away from the true solution (that jointly minimizes PDE residual and BC/IC loss). This analysis also provides guidance for a more informed choice of activation function and initialization for effective PINN training.

We then further provided theoretical results on the benefit of learning in sinusoidal space with PINNs. The sinusoidal mapping is a convenient and effective method to initialize PINN with appropriate input gradient distribution, which helps *sf*-PINNs to overcome the trainability issues encountered by a typical PINN. This is then empirically validated as we demonstrated consistent improvements across different baseline PINN settings for a whole range of forward and inverse modelling problems spanning multiple physics domains. In addition, we made the following observations for *sf*-PINNs.

Firstly, while several works earlier demonstrated the superior performance of Fourier feature mapping relative to standard spatial-temporal inputs, we show in this work that the use of *sin* alone in PINNs can already yield many of the benefits from full Fourier features mapping. We believe it is not so much the use of Fourier features per se that yields the improvement, but the improved initial input gradient properties as outlined in Section IV. Hence, the exact composition of the sinusoidal feature map is less impactful than that they are incorporated in place of the *tanh* function in the input layer. Similarly, while some studies use *sin* activation in all hidden layers, results here suggest that the sinusoidal mappings in the input layer is most important, and we observe that *sf*-PINNs with different activation (*tanh*, *sin*) and initialization (*Xavier*, *He*) schemes used in subsequent hidden layers can produce similar performance. Secondly, through extensive experimental studies, we demonstrate that fine-tuning of initial input gradients through $\sigma$ can lead to significant improvement in *sf*-PINN training results. While this can lead to extra effort, a properly tuned $\sigma$ can also reduce the need for sophisticated tuning of the loss composition required for typical PINNs as discussed in other prior works [34]–[39]. This suggests the use of *sf*-PINNs as an alternate route to addressing some of the PINN training challenges previously identified.

PINNs have shown tremendous potential for being a unifying

AI framework that could assimilate physics theory and measurement data. However, it is yet to become feasible for broad science and engineering applications due to expensive computational cost and training challenges. This can be particularly prohibitive for application to more complex problems. Hence, it is important for future research to take into account fundamental aspects as covered in this work for insights to further improve the efficiency and efficacy of PINN methods.